\documentclass[letterpaper]{article} % DO NOT CHANGE THIS
\usepackage{aaai25}  % DO NOT CHANGE THIS
\usepackage{times}  % DO NOT CHANGE THIS
\usepackage{helvet}  % DO NOT CHANGE THIS
\usepackage{courier}  % DO NOT CHANGE THIS
\usepackage[hyphens]{url}  % DO NOT CHANGE THIS
\usepackage{graphicx} % DO NOT CHANGE THIS
\usepackage{natbib}  % DO NOT CHANGE THIS AND DO NOT ADD ANY OPTIONS TO IT
\usepackage{caption} % DO NOT CHANGE THIS AND DO NOT ADD ANY OPTIONS TO IT
\usepackage{algorithm}
\usepackage{algorithmic}

\usepackage{amsmath,amsfonts,bm}

\def\eqref#1{equation~\ref{#1}}
\def\1{\bm{1}}

\def\va{{\bm{a}}}
\def\vb{{\bm{b}}}

\def\vd{{\bm{d}}}

\def\vx{{\bm{x}}}

\def\mG{{\bm{G}}}

\DeclareMathAlphabet{\mathsfit}{\encodingdefault}{\sfdefault}{m}{sl}
\SetMathAlphabet{\mathsfit}{bold}{\encodingdefault}{\sfdefault}{bx}{n}

\def\sX{{\mathbb{X}}}

\def\emG{{G}}

\newcommand{\normltwo}{L^2}
\newcommand{\normlp}{L^p}

\DeclareMathOperator*{\argmin}{arg\,min}

\usepackage{multirow}

\usepackage{newfloat}
\usepackage{listings}
\DeclareCaptionStyle{ruled}{labelfont=normalfont,labelsep=colon,strut=off} % DO NOT CHANGE THIS
\floatstyle{ruled}
\newfloat{listing}{tb}{lst}{}
\floatname{listing}{Listing}
\title{Functional Connectomes of Neural Networks}
\author{
    Tananun Songdechakraiwut, Yutong Wu
}
\affiliations{
    Department of Computer Science, Duke University\\
    t.song@duke.edu, mason.wu@duke.edu
}

\begin{document}

\maketitle

\begin{abstract}
The human brain is a complex system, and understanding its mechanisms has been a long-standing challenge in neuroscience. The study of the functional connectome, which maps the functional connections between different brain regions, has provided valuable insights through various advanced analysis techniques developed over the years. Similarly, neural networks, inspired by the brain's architecture, have achieved notable success in diverse applications but are often noted for their lack of interpretability. In this paper, we propose a novel approach that bridges neural networks and human brain functions by leveraging brain-inspired techniques. Our approach, grounded in the insights from the functional connectome, offers scalable ways to characterize topology of large neural networks using stable statistical and machine learning techniques. Our empirical analysis demonstrates its capability to enhance the interpretability of neural networks, providing a deeper understanding of their underlying mechanisms.
\end{abstract}


\begin{links}
    \link{Code}{https://github.com/masonwu11/topo-fcnn}
    % \link{Datasets}{https://aaai.org/example/datasets}
    % \link{Extended version}{https://aaai.org/example/extended-version}
    \link{Supplementary material} Available as ancillary files
\end{links}

\section{Introduction}

The human brain is an incredibly complex system, and understanding its intricate workings has been a long-standing challenge in neuroscience. One well-established approach to gaining deeper insights into the brain's underlying mechanisms is through the study of the functional connectome, which maps the functional connections between regions of brain networks and reflects the brain's dynamic network of interactions \cite{bullmore2009complex}. In recent decades, successful findings have emerged from this field, thanks to the development of a wide array of analysis techniques.

Artificial neural networks, inspired by the architecture and functioning of the human brain, have achieved remarkable success in applications ranging from image recognition \cite{he2016deep} to natural language processing \cite{vaswani2017attention}. However, despite these successes, neural networks are often considered black box models due to their lack of interpretability and the difficulty in understanding the underlying mechanisms driving their performance. Neural networks have predesigned architectures with preconfigured weights connecting neurons, analogous to how physically connected brain regions with white matter fibers provide structural information measured by diffusion MRI. Similarly, functional connections between distant neurons resemble how functional MRI measures functional connectivity between brain regions that may not have direct neuroanatomical connections. These connections in the brain give rise to coordinated activity patterns crucial for cognitive processes \cite{honey2009predicting}. Given that neural networks are simplified artificial versions of brain functions, it stands to reason that insights from the functional connectome could be leveraged to enhance our understanding, interpretability, and analysis of these networks, potentially leading to the development of more transparent and efficient models.

However, analyzing functional connectomes is inherently challenging due to the need to extract subtle topological patterns from noisy, complete graphs. Therefore, typical workflows apply a threshold to obtain a sparser graph with a clearer structure before applying techniques from graph theory \cite{bullmore2009complex}. Graph theory has played a crucial role in functional connectome research; however, prior analyses utilizing graph theory have primarily focused on pairwise dyadic relationships, often at a fixed spatial threshold. This approach, centered on dyads, limits the neural structures and functions that graph theory can investigate. Given that a neural network processes information not only based on local neurons of a subnetwork but also across the entire network, from input to output layers, a more comprehensive understanding requires a shift in perspective--from pairwise interactions to capturing higher-order relations (topology) across the full range of spatial resolutions.

\begin{figure*}[t]
\centering
\includegraphics[width=\textwidth]{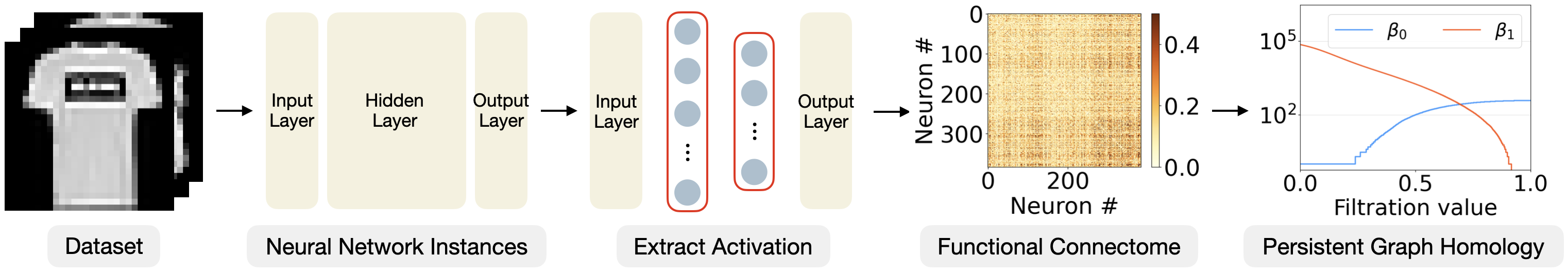} % Reduce the figure size so that it is slightly narrower than the column.
\caption{A schematic for extracting persistent graph homology, representing the topology of neural-network-derived functional connectomes.}
\label{fig:schematic}
\end{figure*}

Persistent homology \cite{edelsbrunner2022computational}, an algebraic topology technique, has emerged as a promising tool for understanding and quantifying the topology of the human brain \cite{sizemore2018cliques,songdechakraiwut2021topological}. Recently, there have been increasing attempts to apply persistent homology to study the interpretability of deep learning. These studies suggest that persistent homology can extract high-order topological information to interpret neural networks, but challenges remain. In particular, current methods often focus solely on the network's structural information, without incorporating data-driven tasks \cite{rieck2018neural,watanabe2022topological}, and are typically limited to models with a small number of neurons \cite{naitzat2020topology,zhang2023functional}. Notably, \citeauthor{zhang2023functional} has applied persistent homology to study the functional connectivities of neural networks. However, their approach is still limited by the cubic time complexity of persistent homology techniques \cite{otter2017roadmap}, making them computationally infeasible for large, complete graphs, and relies heavily on approximation solutions. The authors addressed this challenge by thresholding functional connectivities at a predefined threshold, which limits the spatial resolution of their analysis to a range of pre-determined values, rather than capturing the entire global mechanisms of neural network functions. Importantly, varying threshold values can significantly alter the graph structure, affecting its robustness and sensitivity to signals, and potentially influencing the study’s final outcome.

Persistent graph homology \cite{songdechakraiwut2023topological}, a recent persistent homology advancement inspired by the human brain, captures interpretable topological invariants of great interest in functional brain networks, including connected components and cycles, across the \emph{entire} spatial scale of a graph. Connected components characterize the network shape \cite{tewarie2015minimum}, while cycles represent higher-order interactions potentially associated with information propagation and feedback control \cite{kwon2007analysis}. The computability of persistent graph homology enables the analysis of intricate details in large-scale, complex brain networks. Motivated by this, we adopt a brain-inspired graph homology perspective to overcome computational limitations and fully exploit the potential of topological analyses on large-scale neural network functions.

In this paper, we propose a novel analysis framework that bridges neural networks and human brain functions by leveraging brain-inspired techniques from functional MRI and persistent graph homology. Our framework, grounded in insights from the functional connectome, addresses the aforementioned challenges and offers \emph{scalable} methods to characterize the topology of functional mechanisms of \emph{large} neural networks using \emph{stable} statistical and machine learning techniques. We define functional connectomes, a representation used to describe functions in neural networks, without the need for predefined threshold values, thereby significantly improving the scientific rigor of the analysis. We also demonstrate the computability and efficiency of statistics for such representations via analytic forms. Specifically, we present closed-form computations for the Wasserstein distance, Wasserstein statistics--including barycenter (average) and variance--and the Wasserstein gradient, enabling the realization of the robustness benefits of true Wasserstein distance, grounded in the central stability theorem \cite{skraba2020wasserstein}. Utilizing these computable statistics naturally leads to the development of a centroid-based clustering strategy, where the Wasserstein barycenter serves as the topological centroid. We conducted extensive experiments using this method to validate that our framework can indeed enhance our ability to discern and interpret the complex structure of neural network functions, providing new avenues for both analysis and application.

\section{Functional Connectome Framework}

Figure \ref{fig:schematic} shows a schematic of the functional connectivity analysis framework, described in the following.

\subsection{Functions of Neural Networks}
\label{subsec:functions}

Given a collection of data samples, we partition it into two separate datasets: a training dataset and a functional dataset. The training dataset is utilized via $k$-fold cross-validation to determine a set of optimal hyperparameter values through grid search. Using these optimal values, we train a well-generalized neural network on the entire training data. Subsequently, the functional dataset is fed into the fully-trained neural network, resulting from the aforementioned training process, to investigate the information propagation of new, unknown data through it.

Without loss of generality, we will assume that the neural network architecture of interest is a feedforward network. Given a fully-trained feedforward network with $M$ hidden neurons and a functional dataset denoted by $\sX=\{\vx^{(1)},\vx^{(2)},\ldots,\vx^{(N)}\}$, each data point $\vx^{(i)} \in \sX$ is inputted into the feedforward network and processed through a series of neurons. Each $j$-th neuron uses an affine transformation followed by an activation function to produce its output $a_{ij}$, representing the information processing for $\vx^{(i)}$. By processing the entire functional dataset, we concatenate the $j$-th neuron outputs for all the data points into a \emph{functional vector} $\va_j=(a_{1j},a_{2j},\ldots,a_{Nj})$, representing the functional signal in a neural network, similar to the functions of connectome circuits studied in functional MRI.

When two functional vectors from a pair of neurons are statistically dependent, they exhibit functional synergy. To quantify this synergy, several methods can be used to measure the statistical dependency between the two vectors, such as Pearson correlation, partial correlation, Spearman's rank correlation, and mutual information \cite{fornito2016fundamentals}. In macroscale connectomics, Pearson correlation, which captures linear relationships between different brain regions, is widely used for computing functional connectomes of the human brain. It is straightforward to interpret, with values ranging from -1 to 1, where values further from 0 indicate stronger functional connections. Pearson correlation effectively detects synchrony between brain regions, a common feature of neural activity. For similar reasons, we will use Pearson correlation in this paper to analyze neural networks. Additionally, Pearson correlation's computational efficiency makes it feasible to apply to large neural network models that involve deep layers and many neurons. Formally, Pearson correlation between two functional vectors $\va_j=(a_{1j},a_{2j},...,a_{Nj})$ and $\va_k=(a_{1k},a_{2k},...,a_{Nk})$ is defined as
\begin{equation*}
    \rho_{jk} = \frac{\sum_{i=1}^N(a_{ij} - \overline{\va}_j)(a_{ik} - \overline{\va}_k)}{\sqrt{\sum_{i=1}^N(a_{ij} - \overline{\va}_j)^2} \sqrt{\sum_{i=1}^N(a_{ik} - \overline{\va}_k)^2}},
\end{equation*}
where $\overline{\va}_j=\sum_{i=1}^N a_{ij} \big/ N$ and $\overline{\va}_k=\sum_{i=1}^N a_{ik} \big/ N$.

In studies of the connectome of the human brain, a connectome is typically represented as a graph, comprising brain regions of interest as nodes, and pairwise correlations as edge weights \cite{fornito2016fundamentals}. There is no universally accepted standard for whether the sign of the correlations should be preserved. Generally, choosing the absolute value of negative correlations highlights the strength of connectivity without regard to its direction. This can be particularly useful in analyses where the primary interest is in the magnitude of interactions between brain regions, regardless of whether they are positively or negatively correlated. Additionally, Pearson correlation between the same region is always 1, and thus is typically excluded from analyses, resulting in a brain graph with no self-loops. For similar reasons, we will follow the same well-established procedure applied to the case of a connectome of a neural network. Formally, given the correlations between every pair of hidden neurons, we define an $M$-by-$M$ weighted adjacency matrix $\mG$ whose $jk$-th entry is given as
\begin{equation*}
\emG_{jk} =
    \begin{cases}
    \lvert \rho_{jk} \rvert & \text{if $j \neq k$}; \\
    0 & \text{otherwise}.
    \end{cases}
\end{equation*}
We will call this matrix a \emph{functional connectome} of a neural network.

\subsection{Persistent Graph Homology}

The functional connectome is a very dense matrix, typically representing a \emph{complete} graph. Therefore, typical workflows \cite{bullmore2009complex} apply a threshold to the correlation values for two main reasons: to obtain a sparser graph with a more apparent structure and to reduce the time complexity of computational methods. In particular, methods in persistent homology have cubic time complexity \cite{otter2017roadmap}, making them computationally infeasible for large, complete graphs. This requires iterative, approximation solutions, which introduce numerical errors and reduce the signal-to-noise ratio. Additionally, varying different threshold values significantly alters the graph structure, potentially affecting the study's final outcome. Importantly, pairwise dyadic correlations at a fixed spatial threshold constrain the neural structures and functions that can be investigated. Given that a neural network processes information not only based on local neurons of a subnetwork but also the entire neural network from input to output layers, a more comprehensive understanding requires a shift in perspective from pairwise interactions to capturing higher-order relations across the entire range of spatial resolutions.

In this work, we address these challenges by leveraging brain-inspired \emph{persistent graph homology}, a \emph{scalable} topological-learning paradigm that enables analyses of large-scale functional connectomes \emph{without} approximation \cite{songdechakraiwut2023wasserstein}. It has emerged as a promising tool for understanding, characterizing, and quantifying human connectomes \cite{songdechakraiwut2022fast}. Persistent graph homology describes interpretable topological invariants, including connected components (0th homology group) and independent cycles (1st homology group or cycle rank), across the \emph{entire} spatial scale of a graph.

Formally, given a functional connectome $\mG$, we define a binary graph $\mG_\epsilon$ with the same set of neurons by thresholding the edge correlations so that an edge between neurons $j$ and $k$ exists if $\rho_{jk} > \epsilon$. As $\epsilon$ increases, more edges are removed from the functional connectome $\mG$. Thus, we have a filtration \cite{lee2012persistent}:
$
\mG_{\epsilon_0} \supseteq \mG_{\epsilon_1} \supseteq \cdots \supseteq \mG_{\epsilon_k} ,
$
where $\epsilon_0 \leq \epsilon_1 \leq \cdots \leq \epsilon_k$ are called filtration values.
Persistent homology tracks the birth and death of the topological invariants over these filtration values $\epsilon$. A topological invariant born at filtration $b_j$ and persisting up to filtration $d_j$ is represented by a point $(b_j, d_j)$ in a 2D plane. The set of all such points ${(b_j, d_j)}$ is called a \emph{persistence diagram} \cite{edelsbrunner2022computational}. As $\epsilon$ increases, the number of connected components $\beta_0(\mG_{\epsilon})$ increases monotonically, while the number of cycles $\beta_1(\mG_{\epsilon})$ decreases monotonically. Thus, persistent graph homology only needs to track a collection of sorted birth values $B(\mG)$ for the connected components and a collection of sorted death values $D(\mG)$ for the cycles, given as \cite{songdechakraiwut2023topological}
\begin{align*}
    B(\mG) = \{b_j \}_{j=1}^{M-1}, \quad D(\mG)=\{ d_j \}_{j=1}^{1 + M (M - 3)/2}.
\end{align*}

Figure \ref{fig:schematic} illustrates a schematic for extracting persistent graph homology, which represents the topology of functional connectomes derived from neural networks.

\paragraph{Scalability} The set $B(\mG)$ consists of edge correlations found in the maximum spanning tree (MST) of $\mG$. Once $B(\mG)$ is determined, the set $D(\mG)$ is derived from the edge correlations that are not part of the MST. Therefore, both $B(\mG)$ and $D(\mG)$ can be computed very efficiently in $O(n \log n)$ time, where $n$ is the number of edges in the connectome graph.

\subsection{Persistence Statistics}

Distances are fundamental in statistics because they provide a way to measure how much individual data points vary, enabling the calculation of central tendency, dispersion, and overall data behavior. The Wasserstein distance is a prominent measure in persistent homology, associated with the central concept of the stability theorem \cite{skraba2020wasserstein}. In this section, we will elaborate on the high computability of persistent-graph-homology-based Wasserstein distance and how it results in defining essential statistics such as the mean and variance, among others, which have potential in neural network interpretation.

Specifically, the Wasserstein distance between sets of birth values (or between sets of death values) can be obtained using a closed-form solution. Let $\mG^{(1)}$ and $\mG^{(2)}$ be two given functional connectomes, each having the same number of neurons. Then, the \emph{exact} computation of the $p$-Wasserstein distance is achieved as \cite{songdechakraiwut2023topological}:
\begin{align*}
    W_{p,B}(\mG^{(1)},\mG^{(2)}) &= \Big( \sum_{b \in B(\mG^{(1)})} \lvert b - \tau_0^*(b) \rvert ^p \Big)^{1/p}, \\
    W_{p,D}(\mG^{(1)},\mG^{(2)}) &= \Big( \sum_{d \in D(\mG^{(1)})} \lvert d - \tau_1^*(d) \rvert ^p \Big)^{1/p},
\end{align*}
where $\tau_0^*$ maps the $l$-th smallest birth value in $B(\mG^{(1)})$ to the $l$-th smallest birth value in $B(\mG^{(2)})$, and $\tau_1^*$ maps the $l$-th smallest death value in $D(\mG^{(1)})$ to the $l$-th smallest death value in $D(\mG^{(2)})$, for all $l$.
The exact Wasserstein distances $W_{p,B}$ and $W_{p,D}$ are well-defined because the bijective mappings $\tau_0^*$ and $\tau_1^*$ are well-defined for sets of births and deaths, respectively, with the same cardinality.

Importantly, the analytic expression of the Wasserstein distance above can be equivalently written in a more familiar Euclidean space. To do this, we define a vector of sorted birth values $\vb_{\mG^{(i)}} = (b_{i1}, b_{i2}, ..., b_{i,M-1})$ for the connected components, where $b_{ij} \in B(\mG^{(i)})$ and $b_{ij} \leq b_{i,j+1}$. Similarly, we define a vector of sorted death values $\vd_{\mG^{(i)}} = (d_{i1}, d_{i2}, ..., d_{i, 1 + M (M - 3)/2})$ for the cycles. With these definitions, the $p$-Wasserstein distance can be equivalently expressed as
\begin{align*}
    W_{p,B}(\mG^{(1)},\mG^{(2)}) &= || \vb_{\mG^{(1)}} - \vb_{\mG^{(2)}} ||_p, \\
    W_{p,D}(\mG^{(1)},\mG^{(2)}) &= || \vd_{\mG^{(1)}} - \vd_{\mG^{(2)}} ||_p,
\end{align*}
where $|| \cdot ||_p$ is the $\normlp$ norm. Since $W_{p,B}$ and $W_{p,B}$ are differentiable functions and can be explicitly written down, their gradients are in closed form and can be computed very efficiently.

As is common in machine learning, since we know a computable formula of the Wasserstein distance, and we can take the gradient of that formula efficiently using analytic forms, we can optimize objective functions based on the Wasserstein distance using gradient-based optimization algorithms. For instance, given $N$ functional connectomes $\mG^{(1)},\mG^{(2)},...,\mG^{(N)},$ we can determine a persistent diagram centroid $\overline{\vb}$ (for the connected components) that minimizes the sum of the 2-Wasserstein distances as
\begin{align*}
    \overline{\vb} = \argmin_{\vb_{\overline{\mG}}} \sum_{i=1}^N W_{2,B}(\overline{\mG},\mG^{(i)})
    = \frac{1}{N} \sum_{i=1}^N \vb_{\mG^{(i)}}.
\end{align*}
$\overline{\vb}$ represents the Wasserstein barycenter that quantifies the central tendency of the functional connectomes.
Likewise, the variability around the barycenter can be determined as
\begin{align*}
    s^2_{\vb} = \frac{1}{N} \sum_{i=1}^N (\vb_{\mG^{(i)}} - \overline{\vb}).
\end{align*}
$s^2_{\vb}$ represents the Wasserstein variance. In a similar manner, the Wasserstein mean $\overline{\vd}$ and variance $s^2_{\vd}$ for the cycles can be calculated using $W_{p,D}$.
Additionally, other important statistics that measure central tendency, dispersion, shape, and association can also be computed.

\paragraph{Scalability} The computation of the Wasserstein distance is very efficient. By sorting birth and death values and matching them in order, the computational cost for evaluating $W_{p,B}$ and $W_{p,D}$ is $O(n \log n)$, where $n$ is the number of edges in connectome graphs.

\section{Connectome Analysis of Neural Networks}

Drawing inspiration from functional MRI studies of the human brain, we are interested in the functional behavior of neural networks and whether we can characterize them using our proposed functional connectomes and persistent-graph-homology representation. Through two analysis studies, we aim to explore these connections.

In the first study, we will analyze how various popular regularization strategies, including batch normalization, dropout, and $\normltwo$ regularization, influence the overall functional mechanisms and data propagation in neural networks during inference. Regularization affects neural network weights similarly to how various factors influence structural brain networks obtained by diffusion MRI. This, in turn, impacts how functional mechanisms unfold.

In the second study, we will conduct a more fine-grained investigation into how neural networks process different stimuli through functional connectomes. Specifically, we will explore whether there are inherent topological patterns within the functional mechanisms of processing data points from various predefined classes. For example, we will analyze how neural networks process samples from different digit classes (0-9) in the MNIST dataset. This is analogous to how the human brain perceives and processes different visual stimuli, such as recognizing and distinguishing between digits, characterized by human connectomes observed in function MRI studies.

By drawing these comparisons, we aim to gain insights into the similarities between neural networks and human brain functions to better understand and characterize these systems.

\paragraph{Cluster analysis} We are interested in identifying the natural groupings and relationships within the functional mechanism using our proposed persistent-graph-homology framework. Our goal is to determine if this framework can uncover the intrinsic structure of functional connectomes \emph{without} supervision from predefined classes. Supervised learning can alter the original representation through coefficient and weight adjustments, potentially obscuring the underlying patterns and groupings. Clustering, however, can validate predefined classes by revealing if clusters of functional connectomes align well with them. Specifically, connectomes from the same class should be topologically similar and grouped into the same cluster, while connectomes from different classes should be dissimilar and placed in separate clusters. If this alignment occurs, it supports the idea that the functional mechanisms of neural networks are characterized by their topology and that our framework effectively captures these topological signals. Therefore, we will utilize unsupervised clustering to explore the data, grouping similar connectomes based on their subtle topological patterns.

\paragraph{Method comparison} We evaluated the clustering performance of our proposed method, termed \emph{Top}, relative to six other baseline methods. Our Top method utilizes centroid-based clustering that minimizes within-cluster variances based on squared 2-Wasserstein distances $W^2_{2,B} + W^2_{2,D}$ of birth/death values and the Wasserstein barycenter, optimized via Lloyd's algorithm \cite{forgy1965cluster}.

The first baseline method uses $k$-means clustering on the vectorization of entries below the main diagonal of adjacency (Adj) matrices of functional connectomes, grouping the connectomes based on node-by-node geometry. The remaining five methods use persistent homology and involve clustering on conventional Rips-complex persistence diagrams of the 1st homology group, commonly used in the literature. The $k$-medoids algorithm is applied using the following distances and kernels: bottleneck distance (BD), Wasserstein distance (WD), sliced Wasserstein kernel (SWK) \cite{carriere2017sliced}, and heat kernel (HK) \cite{reininghaus2015stable}. Additionally, $k$-means clustering is performed on the persistence image (PI) vectorization \cite{adams2017persistence}. More details on these methods are available in the supplementary material and code.

For all methods, initial clusters are selected at random, and we perform clustering 20 trials and report average clustering performance.

\paragraph{Datasets} We performed our analyses on three datasets: MNIST \cite{lecun1998gradient}, Fashion-MNIST \cite{xiao2017fashion}, and CIFAR-10 \cite{krizhevsky2009learning}. MNIST comprises a collection of grayscale images of handwritten digits, Fashion-MNIST consists of grayscale images of fashion products, and CIFAR-10 includes color images of various animals and objects. Each of these datasets contains 10 predefined classes. More details on the datasets are available in the supplementary material.

\paragraph{Neural network architectures and training} We employed neural network architectures of increasing complexity to match the varying complexities of the three datasets (MNIST, Fashion-MNIST, and CIFAR-10), while also keeping the architectures compact. This approach ensures that we can train well-generalized neural networks, which will then be analyzed for their behavior in our studies, while maintaining simplicity for transparency and interpretability of the analysis results. Additionally, conventional persistent homology methods, which will serve as baseline methods for comparison with our proposed method, have cubic time complexity \cite{otter2017roadmap}, limiting the architecture size to fewer than a few hundred nodes, as demonstrated in the runtime experiment below.

For MNIST, we used a feedforward architecture with two hidden fully-connected layers, with the first and second layers comprising 128 and 64 neurons, respectively. For Fashion-MNIST, we used a similar feedforward architecture, but with the first and second layers comprising 256 and 128 neurons, respectively. For CIFAR-10, we used a convolutional neural network with three VGG blocks \cite{simonyan2015very}, followed by two fully-connected layers, with the first and second layers comprising 256 and 128 neurons, respectively. In all architectures, we applied leaky ReLU activation functions and the stochastic gradient descent optimizer with momentum.

To train neural networks, we randomly partitioned the data points of each dataset into a training dataset and a functional dataset, as explained in Section \ref{subsec:functions}. For each training strategy--namely 1) batch normalization, 2) dropout, 3) $\normltwo$, as well as 4) vanilla (which trains neural networks without any regularization and serves as a control)--we used the training set to optimize and obtain well-generalized neural networks. To account for the stochastic nature of gradient-based optimization initialization, we trained 20 neural networks for each strategy, totaling 80 networks (20 $\times$ 4).
For the MNIST dataset, the average test accuracies are 0.98 across all training strategies. For Fashion-MNIST, the average test accuracies are 0.89 across all training strategies. For CIFAR-10, the average test accuracies are 0.76 across all training strategies.
% For the MNIST dataset, the average test accuracies are 0.98 for vanilla, 0.98 for batch normalization, 0.98 for dropout, and 0.98 for $\normltwo$. For Fashion-MNIST, the average test accuracies are 0.89 for vanilla, 0.89 for batch normalization, 0.89 for dropout, and 0.89 for $\normltwo$. For CIFAR-10, the average test accuracies are 0.72 for vanilla, 0.80 for batch normalization, 0.81 for dropout, and 0.72 for $\normltwo$.

More details on the architecture and hyperparameter tuning are available in the supplementary material and code.

Once the fully-trained neural networks are obtained, the functional dataset is fed into these networks to extract functional connectomes. Two different methods of extraction are employed for the two analysis studies, which will be provided in more detail below for each specific study. Note that only neurons in the hidden fully-connected layers are used to construct the connectomes; neurons in the softmax output layers and convolutional layers are excluded.

\subsection{Study 1}

\begin{figure*}[t]
\centering
\includegraphics[width=\textwidth]{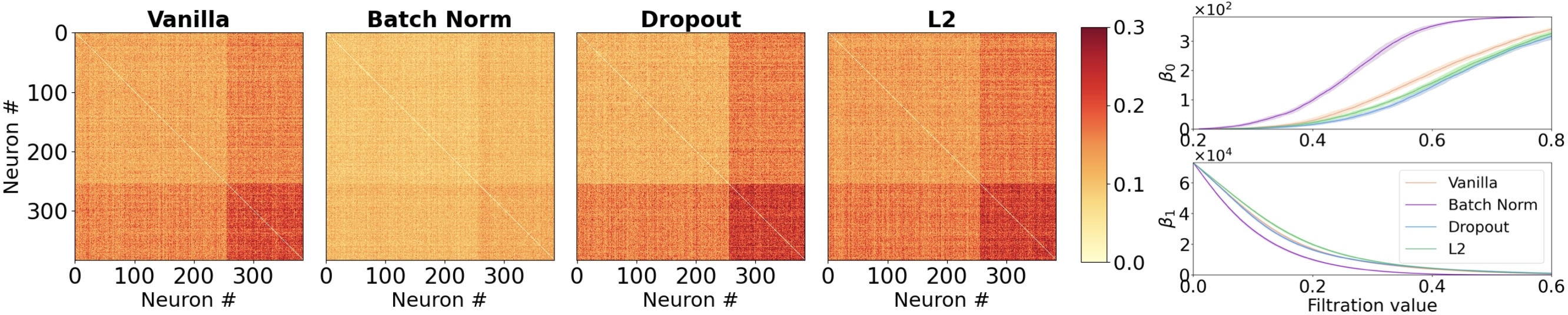}
\caption{Statistics of the functional dataset used in Study 1. 
% The dataset comprises 20 functional connectomes for each of the four training strategies--batch normalization, dropout, $\normltwo$, and vanilla--resulting in a total of 80 connectomes.
Left: Sample means of the functional connectomes, averaged within each training strategy. Right: Persistence diagrams and statistics for each strategy, with thick lines representing Wasserstein barycenters and shaded regions indicating Wasserstein standard deviation.}
\label{fig:connectomes}
\end{figure*}

\begin{table*}[t]
\setlength{\tabcolsep}{.95mm}
\centering
% {\small
\begin{tabular}{ll||ccccccc}
    Dataset & Strategy & BD & WD & SWK & HK & PI & Adj & Top \\
    \hline
    \multirow{3}{4.5em}{MNIST} & All & $0.63 \pm 0.04$ & $0.75$ & \bm{$0.85$} & $0.72 \pm 0.01$ & $0.69 \pm 0.01$ & $0.32 \pm 0.05$ & $0.78 \pm 0.11$ \\
    & Batch Norm vs. Vanilla & \bm{$1.00$} & \bm{$1.00$} & \bm{$1.00$} & $0.93$ & $0.93$ & $0.55 \pm 0.04$ & \bm{$1.00$} \\
    & Dropout vs. Vanilla & $0.67 \pm 0.07$ & $0.90$ & $0.95$ & $0.98$ & $0.98$ & $0.54 \pm 0.03$ & \bm{$1.00$} \\
    & $\normltwo$ vs. Vanilla & $0.62 \pm 0.04$ & $0.70$ & $0.70$ & $0.60$ & $0.58$ & $0.55 \pm 0.03$ & \bm{$0.80 \pm 0.03$} \\
    \hline \hline
    \multirow{3}{4.5em}{Fashion-MNIST} & All & $0.68$ & $0.75 \pm 0.01$ & $0.78$ & $0.52 \pm 0.01$ & $0.53 \pm 0.02$ & $0.34 \pm 0.07$ & \bm{$0.87 \pm 0.12$} \\
    & Batch Norm vs. Vanilla & \bm{$1.00$} & \bm{$1.00$} & \bm{$1.00$} & \bm{$1.00$} & $0.98$ & $0.58 \pm 0.07$ & \bm{$1.00$} \\
    & Dropout vs. Vanilla & $0.73$ & $0.95$ & $0.93$ & $0.50$ & $0.50$ & $0.54 \pm 0.04$ & \bm{$0.98$}\\
    & $\normltwo$ vs. Vanilla & $0.83$ & $0.98$ & $0.95$ & $0.53$ & $0.53$ & $0.55 \pm 0.03$ & \bm{$1.00$}\\
    \hline \hline
    \multirow{3}{4.5em}{CIFAR-10} & All & $0.75 \pm 0.01$ & \bm{$0.98$} & $0.96$ & $0.81$ & $0.58 \pm 0.02$ & $0.51 \pm 0.11$ & $0.88 \pm 0.11$ \\
    & Batch Norm vs. Vanilla & \bm{$1.00$} & \bm{$1.00$} & \bm{$1.00$} & \bm{$1.00$} & \bm{$1.00$} & $0.56 \pm 0.09$ & \bm{$1.00$}\\
    & Dropout vs. Vanilla & \bm{$1.00$} & \bm{$1.00$} & \bm{$1.00$} & \bm{$1.00$} & $0.63 \pm 0.07$ & $0.62 \pm 0.17$ & \bm{$1.00$}\\
    & $\normltwo$ vs. Vanilla & $0.64 \pm 0.01$ & \bm{$0.95$} & $0.93$ & $0.68$ & $0.67 \pm 0.01$ & $0.55 \pm 0.04$ & $0.94 \pm 0.01$\\
    \hline
\end{tabular}
% }
\caption{Comparison of clustering performance in Study 1 across different datasets and training strategies, reported as average purity scores $\pm$ standard deviation. The table presents results for clustering \emph{all} strategies together, as well as pairwise clustering analysis of each regularization strategy compared to the control group (vanilla).}
\label{table:study1}
\end{table*}

\begin{table*}[t]
\setlength{\tabcolsep}{1.07mm}
\centering
% {\small
\begin{tabular}{ll||ccccccc}
    Dataset & Strategy & BD & WD & SWK & HK & PI & Adj & Top \\
    \hline
    \multirow{3}{4.5em}{MNIST} & Vanilla & $0.33$ & $0.40$ & $0.44 \pm 0.02$ & $0.36 \pm 0.01$ & $0.36 \pm 0.01$ & $0.21 \pm 0.02$ & \bm{$0.47 \pm 0.02$}\\
    & Batch Norm & $0.36 \pm 0.02$ & \bm{$0.50$} & $0.48 \pm 0.02$ & $0.35$ & $0.34 \pm 0.01$ & $0.22 \pm 0.03$ & $0.46 \pm 0.02$ \\
    & Dropout & $0.33 \pm 0.02$ & $0.47$ & $0.46 \pm 0.01$ & $0.34 \pm 0.01$ & $0.34 \pm 0.01$ & $0.18 \pm 0.03$ & \bm{$0.57 \pm 0.02$} \\
    & $\normltwo$ & $0.29 \pm 0.01$ & $0.44 \pm 0.01$ & $0.45 \pm 0.01$ & $0.34 \pm 0.01$ & $0.34 \pm 0.01$ & $0.19 \pm 0.02$ & \bm{$0.48 \pm 0.02$} \\
    \hline \hline
    \multirow{3}{4.5em}{Fashion-MNIST} & Vanilla & $0.39 \pm 0.01$ & $0.62 \pm 0.01$ & \bm{$0.64 \pm 0.02$} & $0.46 \pm 0.02$ & $0.43 \pm 0.01$ & $0.23 \pm 0.03$ & $0.53 \pm 0.02$\\
    & Batch Norm & $0.38 \pm 0.02$ & $0.58 \pm 0.02$ & \bm{$0.60 \pm 0.01$} & $0.40 \pm 0.01$ & $0.39 \pm 0.01$ & $0.20 \pm 0.04$ & $0.49 \pm 0.02$\\
    & Dropout & $0.41$ & \bm{$0.60$} & $0.59 \pm 0.04$ & $0.45 \pm 0.02$ & $0.40 \pm 0.01$ & $0.19 \pm 0.03$ & $0.53 \pm 0.03$ \\
    & $\normltwo$ & $0.43$ & $0.54 \pm 0.01$ & \bm{$0.59 \pm 0.01$} & $0.41$ & $0.41 \pm 0.01$ & $0.21 \pm 0.04$ & $0.53 \pm 0.04$\\
    \hline \hline
    \multirow{3}{4.5em}{CIFAR-10} & Vanilla & $0.30 \pm 0.01$ & \bm{$0.56 \pm 0.01$} & $0.55 \pm 0.01$ & $0.41 \pm 0.01$ & $0.40 \pm 0.01$ & $0.15 \pm 0.02$ & $0.51 \pm 0.03$ \\
    & Batch Norm & $0.33 \pm 0.01$ & $0.51 \pm 0.01$ & $0.47$ & $0.35 \pm 0.01$ & $0.35 \pm 0.01$ & $0.16 \pm 0.01$ & \bm{$0.52 \pm 0.02$}\\
    & Dropout & $0.29 \pm 0.01$ & $0.49$ & \bm{$0.51 \pm 0.01$} & $0.37 \pm 0.01$ & $0.37 \pm 0.01$ & $0.26 \pm 0.04$ & $0.50 \pm 0.02$\\
    & $\normltwo$ & $0.35 \pm 0.01$ & $0.59 \pm 0.03$ & \bm{$0.59 \pm 0.01$} & $0.50 \pm 0.01$ & $0.48 \pm 0.01$ & $0.18 \pm 0.03$ & $0.54 \pm 0.03$\\
    \hline
\end{tabular}
% }
\caption{Comparison of clustering performance in Study 2, reported as average purity scores $\pm$ standard deviation.}
\label{table:study2}
\end{table*}

We will analyze the influence of various popular regularization strategies on the \emph{overall} functional mechanisms and data propagation in neural networks during inference. To construct functional connectomes for each dataset, we feed the \emph{entire} corresponding functional dataset to the neural network to extract functional connectomes. As a result, we obtain 80 functional connectomes, with 20 of these connectomes from each training strategy (batch normalization, dropout, $\normltwo$, and vanilla). We will perform cluster analysis on these 80 data points to group them into four clusters.
Figure \ref{fig:connectomes} illustrates the average functional connectomes from training on the Fashion-MNIST dataset using each strategy, along with their corresponding persistence diagrams, which describe the topology.
Additionally, we will cluster each regularization strategy against the control group (i.e., vanilla) to better understand the impact of each regularization method on the neural networks' functional mechanisms. That is, we will cluster 40 data points into two groups.

Since these datasets are balanced, with 20 samples per class, we can evaluate clustering performance using \emph{purity} \cite{manning2008prabhakar}. The purity score ranges from 0 to 1, with 1 indicates perfect clustering alignment.
This evaluation measure is not only transparent and interpretable but also effective for this study, where the number of clusters is small and the cluster sizes are balanced.

Table \ref{table:study1} presents the clustering performance comparison. The findings indicate that regularization strategies, which influence learnable weight adjustment of neural networks, significantly affect data propagation through functional mechanisms, similar to neuroanatomy in the human nervous system \cite{sporns2016networks}. Specifically, the clustering performance shows high purity scores, which suggest substantial differences between each regularization strategy vs. the control, highlighting the notable impact of these techniques. Furthermore, different regularization strategies lead to distinct functional mechanisms, resulting in high purity scores in cluster analysis for all four training strategies. Overall, topological signals, as measured by persistent homology methods, prove to be an effective means of characterizing neural network functions. In most settings, the proposed Top method outperforms other baselines.

\subsection{Study 2}

We will explore how fully-trained neural networks process different stimuli using functional connectomes. We construct these connectomes as follows. For each dataset, we partition the data into collections where each collection contains data points from a specific predefined class. For instance, in the MNIST dataset, we create 10 collections, each corresponding to a digit class (0-9). We then feed each collection into the fully-trained neural network to extract the functional connectomes for that particular class. For each training strategy, we obtain 20 functional connectomes per class, resulting in a total of 200 functional connectomes (20 $\times$ 10). We will perform cluster analysis on these 200 data points to group them into ten clusters.

As in Study 1, these datasets in Study 2 are balanced so we will also use the purity score to evaluate clustering performance.

Table \ref{table:study2} displays the clustering performance comparison. Topological methods, including WD, SWK and Top, effectively capture the functions distinct to each predefined class. They are the best performers that achieve purity scores between 0.5 and 0.6 in \emph{unsupervised} settings, which are significantly better than the 0.1 score expected if clustering was made randomly. These findings show that samples from each class are processed through distinct functional mechanisms. These phenomena are observed in all training strategies. This is similar to how the human brain uses specialized neural mechanisms to process different types of stimuli, ensuring efficient and effective interpretation of diverse information \cite{kandel2000principles}.

\subsection{Runtime Experiment}

\begin{figure}[t]
\centering
\includegraphics[width=\columnwidth]{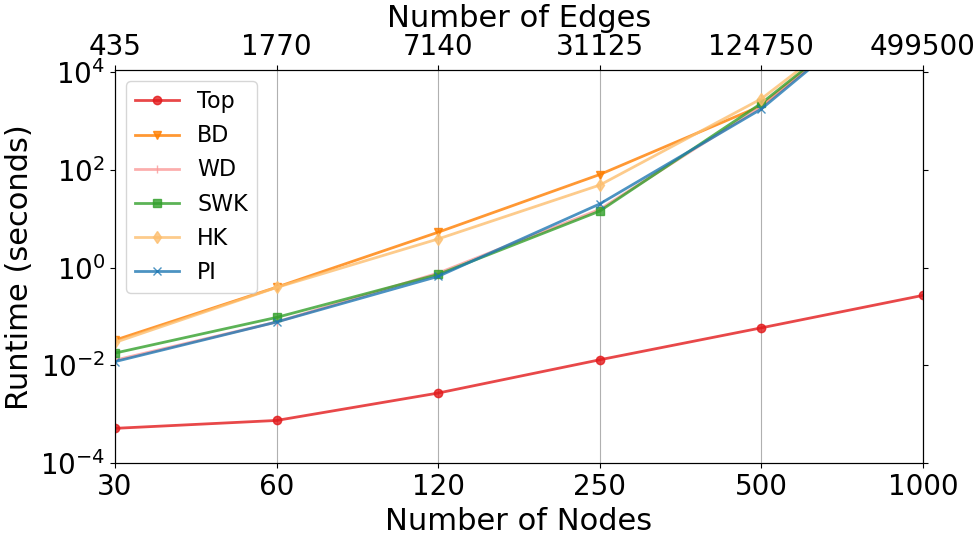}
\caption{Average runtime of each method for computing topological distance or kernel between two complete graphs. The graphs were generated using a modular network approach (details in supplementary material, with code provided). The runtime is plotted against the network size, represented by the number of nodes and edges.}
\label{fig:runtime}
\end{figure}

All topological methods used in the studies were evaluated through runtime experiments. These methods were executed on an Apple M1 Pro CPU with 16 GB of unified RAM. Figure \ref{fig:runtime} shows the plot of runtime vs. input size. The results clearly indicate that all five persistent homology-based distances and kernels (BD, WD, SWK, HK, and PI) are limited to handling dense graphs with only a few hundred nodes, highlighting the current scaling limitations of persistent homology embedding methods and their heavy reliance on approximation solutions. In contrast, Top can compute the exact Wasserstein distance between graphs with thousands of nodes and millions of edges in about one second. This computational efficiency makes Top practical for large-scale analyses of neural networks, which cannot be effectively analyzed using methods based on conventional persistence diagrams.

\paragraph{Potential Impact}

Our approach for characterizing functional connectomes in neural networks is effective, computable, and scalable, potentially impacting the analysis of large neural architectures. By integrating neural network interpretability with human brain function insights, our framework opens up opportunities to leverage established techniques from functional MRI analysis.
From a statistical learning perspective, our persistence statistics provide a robust basis for hypothesis testing and permutation tests, enhancing statistical rigor. Their linear-logarithmic efficiency supports large-scale neural network applications. Additionally, the gradient computability of the Wasserstein distance aids in designing advanced machine learning algorithms through gradient descent optimization.

Our studies on convolutional neural networks for the CIFAR-10 dataset demonstrate the effectiveness of focusing on subnetworks within the last few fully-connected layers, enabling topological analysis of more targeted functional mechanisms. This approach could be particularly effective for complex, deep neural networks, including those with multiple heads. While primarily focused on feedforward architectures, our method can also be extended to convolutional layers and recurrent networks.

\paragraph{Limitation}
Persistent graph homology is limited to the topological invariants of connected components and cycles. These two features, however, play a dominant role in topological analyses. For example, they are widely utilized in the brain network community \cite{bullmore2009complex,honey2007network}, and cycles, in particular, have been increasingly reported as the most discriminative topological feature in brain networks \cite{sizemore2018cliques}, galaxy organization \cite{biagetti2021persistence}, and protein structure \cite{xia2014persistent}. In contrast, the assessment of higher-order features beyond cycles offers limited practical value due to their relative rarity, interpretive challenges, and consequent minimal discriminative power \cite{biagetti2021persistence,sizemore2018cliques,songdechakraiwut2020dynamic}.

\bibliography{aaai25}

\begin{thebibliography}{35}
\providecommand{\natexlab}[1]{#1}

\bibitem[{Adams et~al.(2017)Adams, Emerson, Kirby, Neville, Peterson, Shipman, Chepushtanova, Hanson, Motta, and Ziegelmeier}]{adams2017persistence}
Adams, H.; Emerson, T.; Kirby, M.; Neville, R.; Peterson, C.; Shipman, P.; Chepushtanova, S.; Hanson, E.; Motta, F.; and Ziegelmeier, L. 2017.
\newblock Persistence images: a stable vector representation of persistent homology.
\newblock \emph{Journal of Machine Learning Research}, 18(8): 1--35.

\bibitem[{Biagetti, Cole, and Shiu(2021)}]{biagetti2021persistence}
Biagetti, M.; Cole, A.; and Shiu, G. 2021.
\newblock The persistence of large scale structures. {Part I. Primordial non-Gaussianity}.
\newblock \emph{Journal of Cosmology and Astroparticle Physics}, 2021(04): 061.

\bibitem[{Bullmore and Sporns(2009)}]{bullmore2009complex}
Bullmore, E.; and Sporns, O. 2009.
\newblock Complex brain networks: graph theoretical analysis of structural and functional systems.
\newblock \emph{Nature Reviews Neuroscience}, 10(3): 186--198.

\bibitem[{Carriere, Cuturi, and Oudot(2017)}]{carriere2017sliced}
Carriere, M.; Cuturi, M.; and Oudot, S. 2017.
\newblock Sliced Wasserstein kernel for persistence diagrams.
\newblock In \emph{International Conference on Machine Learning (ICML)}, 664--673.

\bibitem[{Edelsbrunner and Harer(2022)}]{edelsbrunner2022computational}
Edelsbrunner, H.; and Harer, J.~L. 2022.
\newblock \emph{Computational Topology: An Introduction}.
\newblock American Mathematical Society.

\bibitem[{Forgy(1965)}]{forgy1965cluster}
Forgy, E.~W. 1965.
\newblock Cluster analysis of multivariate data: efficiency versus interpretability of classifications.
\newblock \emph{Biometrics}, 21: 768--769.

\bibitem[{Fornito, Zalesky, and Bullmore(2016)}]{fornito2016fundamentals}
Fornito, A.; Zalesky, A.; and Bullmore, E. 2016.
\newblock \emph{Fundamentals of Brain Network Analysis}.
\newblock Academic press.

\bibitem[{He et~al.(2016)He, Zhang, Ren, and Sun}]{he2016deep}
He, K.; Zhang, X.; Ren, S.; and Sun, J. 2016.
\newblock Deep residual learning for image recognition.
\newblock In \emph{Proceedings of the IEEE Conference on Computer Vision and Pattern Recognition (CVPR)}, 770--778.

\bibitem[{Honey et~al.(2007)Honey, K{\"o}tter, Breakspear, and Sporns}]{honey2007network}
Honey, C.~J.; K{\"o}tter, R.; Breakspear, M.; and Sporns, O. 2007.
\newblock Network structure of cerebral cortex shapes functional connectivity on multiple time scales.
\newblock \emph{Proceedings of the National Academy of Sciences}, 104(24): 10240--10245.

\bibitem[{Honey et~al.(2009)Honey, Sporns, Cammoun, Gigandet, Thiran, Meuli, and Hagmann}]{honey2009predicting}
Honey, C.~J.; Sporns, O.; Cammoun, L.; Gigandet, X.; Thiran, J.-P.; Meuli, R.; and Hagmann, P. 2009.
\newblock Predicting human resting-state functional connectivity from structural connectivity.
\newblock \emph{Proceedings of the National Academy of Sciences}, 106(6): 2035--2040.

\bibitem[{Kandel et~al.(2000)Kandel, Schwartz, Jessell, Siegelbaum, Hudspeth, Mack et~al.}]{kandel2000principles}
Kandel, E.~R.; Schwartz, J.~H.; Jessell, T.~M.; Siegelbaum, S.; Hudspeth, A.~J.; Mack, S.; et~al. 2000.
\newblock \emph{Principles of Neural Science}, volume~4.
\newblock McGraw-hill New York.

\bibitem[{Krizhevsky, Hinton et~al.(2009)}]{krizhevsky2009learning}
Krizhevsky, A.; Hinton, G.; et~al. 2009.
\newblock Learning multiple layers of features from tiny images.

\bibitem[{Kwon and Cho(2007)}]{kwon2007analysis}
Kwon, Y.-K.; and Cho, K.-H. 2007.
\newblock Analysis of feedback loops and robustness in network evolution based on Boolean models.
\newblock \emph{BMC Bioinformatics}, 8.

\bibitem[{LeCun et~al.(1998)LeCun, Bottou, Bengio, and Haffner}]{lecun1998gradient}
LeCun, Y.; Bottou, L.; Bengio, Y.; and Haffner, P. 1998.
\newblock Gradient-based learning applied to document recognition.
\newblock \emph{Proceedings of the IEEE}, 86(11): 2278--2324.

\bibitem[{Lee et~al.(2012)Lee, Kang, Chung, Kim, and Lee}]{lee2012persistent}
Lee, H.; Kang, H.; Chung, M.~K.; Kim, B.-N.; and Lee, D.~S. 2012.
\newblock Persistent brain network homology from the perspective of dendrogram.
\newblock \emph{IEEE Transactions on Medical Imaging}, 31(12): 2267--2277.

\bibitem[{Manning, Raghavan, and Schütze(2008)}]{manning2008prabhakar}
Manning, C.~D.; Raghavan, P.; and Schütze, H. 2008.
\newblock \emph{Introduction to Information Retrieval}.
\newblock Cambridge University Press.

\bibitem[{Naitzat, Zhitnikov, and Lim(2020)}]{naitzat2020topology}
Naitzat, G.; Zhitnikov, A.; and Lim, L.-H. 2020.
\newblock Topology of deep neural networks.
\newblock \emph{Journal of Machine Learning Research}, 21(184): 1--40.

\bibitem[{Otter et~al.(2017)Otter, Porter, Tillmann, Grindrod, and Harrington}]{otter2017roadmap}
Otter, N.; Porter, M.~A.; Tillmann, U.; Grindrod, P.; and Harrington, H.~A. 2017.
\newblock A roadmap for the computation of persistent homology.
\newblock \emph{EPJ Data Science}, 6: 1--38.

\bibitem[{Reininghaus et~al.(2015)Reininghaus, Huber, Bauer, and Kwitt}]{reininghaus2015stable}
Reininghaus, J.; Huber, S.; Bauer, U.; and Kwitt, R. 2015.
\newblock A stable multi-scale kernel for topological machine learning.
\newblock In \emph{Proceedings of the IEEE Conference on Computer Vision and Pattern Recognition (CVPR)}, 4741--4748.

\bibitem[{Rieck et~al.(2019)Rieck, Togninalli, Bock, Moor, Horn, Gumbsch, and Borgwardt}]{rieck2018neural}
Rieck, B.; Togninalli, M.; Bock, C.; Moor, M.; Horn, M.; Gumbsch, T.; and Borgwardt, K. 2019.
\newblock Neural persistence: a complexity measure for deep neural networks using algebraic topology.
\newblock In \emph{International Conference on Learning Representations (ICLR)}.

\bibitem[{Simonyan and Zisserman(2015)}]{simonyan2015very}
Simonyan, K.; and Zisserman, A. 2015.
\newblock Very deep convolutional networks for large-scale image recognition.
\newblock In \emph{International Conference on Learning Representations (ICLR)}.

\bibitem[{Sizemore et~al.(2018)Sizemore, Giusti, Kahn, Vettel, Betzel, and Bassett}]{sizemore2018cliques}
Sizemore, A.~E.; Giusti, C.; Kahn, A.; Vettel, J.~M.; Betzel, R.~F.; and Bassett, D.~S. 2018.
\newblock Cliques and cavities in the human connectome.
\newblock \emph{Journal of Computational Neuroscience}, 44: 115--145.

\bibitem[{Skraba and Turner(2023)}]{skraba2020wasserstein}
Skraba, P.; and Turner, K. 2023.
\newblock Wasserstein stability for persistence diagrams.
\newblock \emph{arXiv preprint arXiv:2006.16824}.

\bibitem[{Songdechakraiwut and Chung(2020)}]{songdechakraiwut2020dynamic}
Songdechakraiwut, T.; and Chung, M.~K. 2020.
\newblock Dynamic topological data analysis for functional brain signals.
\newblock In \emph{IEEE International Symposium on Biomedical Imaging}, 1--4.

\bibitem[{Songdechakraiwut and Chung(2023)}]{songdechakraiwut2023topological}
Songdechakraiwut, T.; and Chung, M.~K. 2023.
\newblock Topological learning for brain networks.
\newblock \emph{The Annals of Applied Statistics}, 17(1): 403.

\bibitem[{Songdechakraiwut et~al.(2023)Songdechakraiwut, Krause, Banks, Nourski, and Van~Veen}]{songdechakraiwut2023wasserstein}
Songdechakraiwut, T.; Krause, B.~M.; Banks, M.~I.; Nourski, K.~V.; and Van~Veen, B.~D. 2023.
\newblock Wasserstein distance-preserving vector space of persistent homology.
\newblock In \emph{International Conference on Medical Image Computing and Computer-Assisted Intervention (MICCAI)}, 277--286.

\bibitem[{Songdechakraiwut et~al.(2022)Songdechakraiwut, Krause, Banks, Nourski, and Veen}]{songdechakraiwut2022fast}
Songdechakraiwut, T.; Krause, B.~M.; Banks, M.~I.; Nourski, K.~V.; and Veen, B. D.~V. 2022.
\newblock Fast topological clustering with Wasserstein distance.
\newblock In \emph{International Conference on Learning Representations (ICLR)}.

\bibitem[{Songdechakraiwut, Shen, and Chung(2021)}]{songdechakraiwut2021topological}
Songdechakraiwut, T.; Shen, L.; and Chung, M. 2021.
\newblock Topological learning and its application to multimodal brain network integration.
\newblock In \emph{International Conference on Medical Image Computing and Computer Assisted Intervention (MICCAI)}, 166--176.

\bibitem[{Sporns(2016)}]{sporns2016networks}
Sporns, O. 2016.
\newblock \emph{Networks of the Brain}.
\newblock MIT press.

\bibitem[{Tewarie et~al.(2015)Tewarie, van Dellen, Hillebrand, and Stam}]{tewarie2015minimum}
Tewarie, P.; van Dellen, E.; Hillebrand, A.; and Stam, C.~J. 2015.
\newblock The minimum spanning tree: an unbiased method for brain network analysis.
\newblock \emph{NeuroImage}, 104: 177--188.

\bibitem[{Vaswani et~al.(2017)Vaswani, Shazeer, Parmar, Uszkoreit, Jones, Gomez, Kaiser, and Polosukhin}]{vaswani2017attention}
Vaswani, A.; Shazeer, N.; Parmar, N.; Uszkoreit, J.; Jones, L.; Gomez, A.~N.; Kaiser, L.~u.; and Polosukhin, I. 2017.
\newblock Attention is all you need.
\newblock In \emph{Advances in Neural Information Processing Systems}, volume~30.

\bibitem[{Watanabe and Yamana(2022)}]{watanabe2022topological}
Watanabe, S.; and Yamana, H. 2022.
\newblock Topological measurement of deep neural networks using persistent homology.
\newblock \emph{Annals of Mathematics and Artificial Intelligence}, 90(1): 75--92.

\bibitem[{Xia and Wei(2014)}]{xia2014persistent}
Xia, K.; and Wei, G.-W. 2014.
\newblock Persistent homology analysis of protein structure, flexibility, and folding.
\newblock \emph{International Journal for Numerical Methods in Biomedical Engineering}, 30(8): 814--844.

\bibitem[{Xiao, Rasul, and Vollgraf(2017)}]{xiao2017fashion}
Xiao, H.; Rasul, K.; and Vollgraf, R. 2017.
\newblock Fashion-{MNIST}: a novel image dataset for benchmarking machine learning algorithms.
\newblock \emph{arXiv preprint arXiv:1708.07747}.

\bibitem[{Zhang et~al.(2023)Zhang, Dong, Zhang, and Lin}]{zhang2023functional}
Zhang, B.; Dong, Z.; Zhang, J.; and Lin, H. 2023.
\newblock Functional network: a novel framework for interpretability of deep neural networks.
\newblock \emph{Neurocomputing}, 519: 94--103.

\end{thebibliography}

\end{document}